\documentclass{article}

\usepackage{arxiv}

\usepackage[utf8]{inputenc} % allow utf-8 input
\usepackage[T1]{fontenc}    % use 8-bit T1 fonts
\usepackage{hyperref}       % hyperlinks
\usepackage{url}            % simple URL typesetting
\usepackage{booktabs}       % professional-quality tables
\usepackage{amsfonts}       % blackboard math symbols
\usepackage{nicefrac}       % compact symbols for 1/2, etc.
\usepackage{microtype}      % microtypography
\usepackage{lipsum}
\usepackage{graphicx}
\usepackage[utf8]{inputenc} % allow utf-8 input
\usepackage[T1]{fontenc}    % use 8-bit T1 fonts
\usepackage{hyperref}       % hyperlinks
\usepackage{url}            % simple URL typesetting
\usepackage{booktabs}       % professional-quality tables
\usepackage{amsfonts}       % blackboard math symbols
\usepackage{nicefrac}       % compact symbols for 1/2, etc.
\usepackage{microtype}      % microtypography
\usepackage{xcolor}         % colors
\usepackage{graphicx}
\usepackage{subfigure}
\usepackage{amsmath}
\usepackage{amsthm}
\usepackage{amssymb}
\usepackage{enumitem}
\usepackage{adjustbox}
\usepackage{wrapfig}
\usepackage[normalem]{ulem}
\usepackage{mathrsfs}
\usepackage{multirow}
\usepackage{svg}
\newtheorem{theorem}{Theorem}

\usepackage{algorithm}
\usepackage{algorithmic}
\theoremstyle{definition}
\newtheorem{definition}[theorem]{Definition}

\newcommand\algorithmicprocedure{\textbf{function}}
\newcommand{\algorithmicendprocedure}{\algorithmicend\ \algorithmicprocedure}
\makeatletter
\newcommand\FUNCTION[3][default]{%
  \ALC@it
  \algorithmicprocedure\ \textsc{#2}(#3)%
  \ALC@com{#1}%
  \begin{ALC@prc}%
}
\newcommand\ENDFUNCTION{%
  \end{ALC@prc}%
  \ifthenelse{\boolean{ALC@noend}}{}{%
    \ALC@it\algorithmicendprocedure
  }%
}
\newenvironment{ALC@prc}{\begin{ALC@g}}{\end{ALC@g}}

\graphicspath{ {./figures/} }

\title{Scalable Neural Symbolic Regression using Control Variables}
\author{
 Xieting Chu \\
 School of Computer Science \\
  GaTech \\
  \texttt{creatixchu@gatech.edu} \\
   \And
 Hongjue Zhao \\
  School of Computing and Data Science \\
  UIUC \\
  \texttt{hongjue2@illinois.edu} \\
  \And
  Enze Xu \\
  Department of Computer Science \\
  William \& Mary \\
  \texttt{exu03@wm.edu} \\
  \And
  Hairong Qi \\
  Department of EECS \\
  University of Tennessee, Knoxville \\
  \texttt{hqi@utk.edu} \\
  \And
  Minghan Chen \\
  Department of Computer Science \\
  Wake Forest University \\
  \texttt{chenm@wfu.edu} \\
  \And
  Huajie Shao \\
 Department of Computer Science \\
  William \& Mary \\
  \texttt{hshao@wm.edu} \\
}

\begin{document}
\maketitle
\begin{abstract}
Symbolic regression (SR) is a powerful technique for discovering the analytical mathematical expression from data, finding various applications in natural sciences due to its good interpretability of results. However, existing methods face scalability issues when dealing with complex equations involving multiple variables. To address this challenge, we propose ScaleSR, a scalable symbolic regression model that leverages control variables to enhance both accuracy and scalability. The core idea is to decompose multi-variable symbolic regression into a set of single-variable SR problems, which are then combined in a bottom-up manner. The proposed method involves a four-step process. First, we learn a data generator from observed data using deep neural networks (DNNs). Second, the data generator is used to generate samples for a certain variable by controlling the input variables. Thirdly, single-variable symbolic regression is applied to estimate the corresponding mathematical expression. Lastly, we repeat steps 2 and 3 by gradually adding variables one by one until completion. We evaluate the performance of our method on multiple benchmark datasets. Experimental results demonstrate that the proposed ScaleSR significantly outperforms state-of-the-art baselines in discovering mathematical expressions with multiple variables. Moreover, it can substantially reduce the search space for symbolic regression. The source code will be made publicly available upon publication.
\end{abstract}

% keywords can be removed

%%%%%%%%% BODY %%%%%%%%%%%%
\section{Introduction}\label{sec:intro}
Symbolic regression (SR) aims to uncover the underlying mathematical expressions from observed data~\cite{la2021contemporary,cranmer2020discovering}. It has been widely used for scientific discovery across various disciplines~\cite{alaa2019demystifying,sun2022symbolic} owing to its ability to learn analytical expressions between the input and output. The implementation of SR involves two steps~\cite{kamienny2022end}. The first step is to predict the skeleton of mathematical expressions based on a pre-defined list of basic operations ($+, -, \times, \div$) and functions ($\sin, \cos, \exp, \log$). For instance, we can identify the skeleton of a symbolic equation as $f(x)=\log{ax}+\sin(bx) + c$. Next, we adopt optimization methods, such as Broyden–Fletcher–Goldfarb–Shanno (BFGS), to estimate the parameters $a,b,c$ in the skeleton. The key challenges of SR lie in: 1) how to improve the accuracy and scalability for multiple input variables, and 2) how to speed up the discovery process.

% interpretable analytical expression

In the past few decades, a plethora of SR methods~\cite{makke2022interpretable} have been developed to discover underlying mathematical equations from data in science and engineering domains. One popular approach among them is genetic programming (GP)~\cite{billard2003statistics,cornforth2012symbolic,quade2016prediction,de2021interaction,arnaldo2014multiple}, which uses evolutionary operations, such as mutation, crossover, and selection, to estimate the symbolic expressions in a tree structure. However, GP would suffer from instability and its inference time is expensive in the context of multiple input variables~\cite{kamienny2022end}. Another method, SINDy~\cite{brunton2016discovering}, adopts sparse linear regression to discover the governing equations of dynamical systems. However, SINDy's performance relies heavily on prior knowledge of a known set of candidate functions, and it is difficult to uncover complex equations from data solely through linear regression. To overcome these limitations, some studies explore deep neural networks-based techniques, such as Deep Symbolic Regression (DSR)~\cite{petersendeep} and Transformer-based pre-training, for symbolic learning. Although these approaches obtain good prediction accuracy, they do not scale well to mathematical equations with multiple variables. Recently, researchers develop Symbolic Physics Learner (SPL), a physics-informed Monte Carlo Tree Search (MCTS) algorithm for symbolic regression. While SPL outperforms most GP-based methods, it still struggles with multiple variables in mathematical expressions. In summary, existing methods suffer from scalability issues when dealing with complex multi-variable equations as they require a much larger search space to identify the combination of different variables. Thus, the question is, how can we reduce the search space of symbolic regression for complex equations involving multiple variables?

\begin{figure*}[!tb]
\begin{center}
 \includegraphics[width=\textwidth]{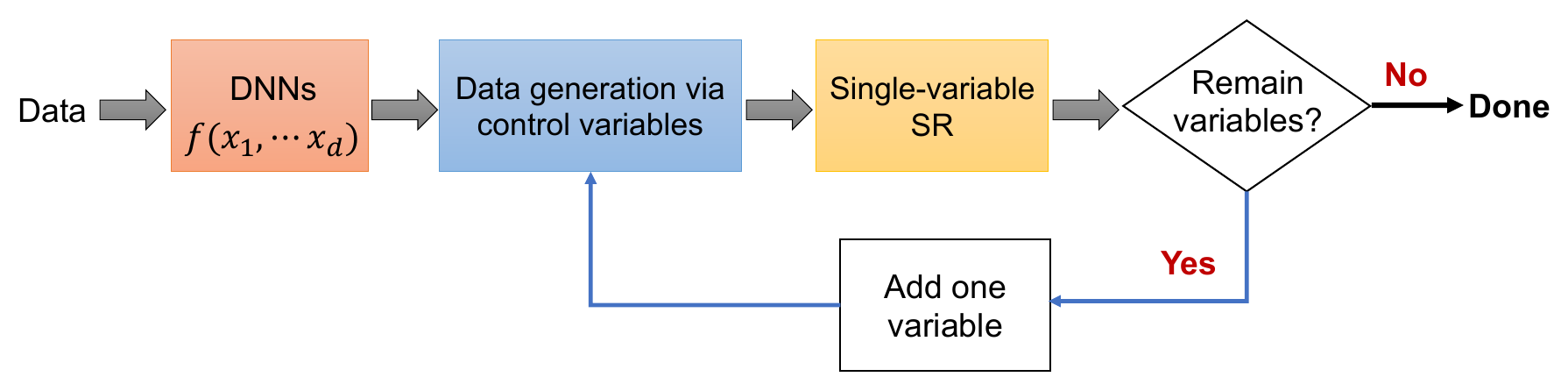}
 \vspace{-0.2in}
\caption{The overall framework of ScaleSR, consisting of three main components: i) learn a data generator using DNNs; ii) generate data for each independent variable via control variables; iii) apply single-variable SR to estimate the mathematical equation for the current independent variable.}\label{fig:framework}
\vspace{-0.2in}
\end{center}
\end{figure*}

In this paper, we propose ScaleSR, a novel neural symbolic regression with control variables to discover analytical expressions from data, as illustrated in Fig.~\ref{fig:framework}. Inspired by divide and conquer~\cite{posner2010divide}, ScaleSR addresses the multi-variable symbolic regression by decomposing it into a set of single-variable SR problems and then combines the estimated symbolic equation for each variable in a bottom-up manner. The proposed method is performed in four steps as follows. 1) We use DNNS to learn a surrogate model to approximate observed data, allowing for generating data for a specific variable. 2) Generate data via control variables. Specifically, we generate data samples for the current independent variable by manipulating the previously learned variables and other control variables. For example, for estimating the symbolic expression of variable $x_i$, we can generate data samples by varying $x_i$ while fixing the other variables. 3) Single-variable symbolic regression is employed to estimate the mathematical expression of the current variable based on the generated data in step 2. Here any symbolic regression models can be inserted into the framework. 4) We gradually add the remaining variables one by one to step 2 and proceed with step 3 until all the variables are covered. Extensive experimental results on multiple SR benchmarks demonstrate the superiority of our ScaleSR over the state-of-the-art methods in discovering complex multi-variable equations. Moreover, the proposed approach is able to discover complex expressions in a reduced search space.

Our main contributions are three-fold: 1) we propose ScaleSR, a simple and effective neural symbolic regression method using control variables; 2) we illustrate that the proposed method exhibits a significant reduction in search space for complex symbolic equations; 3) the evaluation results demonstrate that our method can significantly outperform the baselines in terms of accuracy and inference time.

% The basic idea is that NN serves as a data generator that generates the trajectories of each single variable while keeping the other variables fixed
%%-----overall framework------

\section{Related Work}\label{sec:relatedwork}
% \shao{past tense of verb}
\textbf{GP-based Symbolic Regression.} Genetic Programming (GP) is one of the most popular algorithms for symbolic regression. The basic idea is to adopt the evolutionary operations, including mutation, crossover, and selection, to iteratively estimate the mathematical expressions until the desired accuracy is achieved. As a typical representative, the commercial software Eureqa~\cite{dubvcakova2011eureqa} has been widely used in real-world applications. A recent study~\cite{mundhenk2021symbolic} combined genetic programming with reinforcement learning to enhance performance. While GP yields satisfactory results in many scenarios, it does not scale well to multiple input variables and is highly sensitive to hyperparameters~\cite{petersen2019deep}. 

% \hongjue{Recent years there is also researcher combining genetic programming with deep reinforcement learning so as to leverage each of there strengths\cite{mundhenk2021symbolic}. However, it always fails on Nguyen-12, which we can recover with the success rate 70\% shown in Table.~\ref{tab:comparison_sr}.} 

%%% DNNs based Symbolic regression
\textbf{DNNs-based Symbolic Regression.} Some studies have employed DNN techniques~\cite{landajuela2021discovering, landajuela2022unified, li2022console} to discover symbolic equations from data. Early approaches proposed to replace the activation functions in DNNs with some basic functions like ``$\sin(.)$'', ``$\cos(.)$'', and ``$\exp(.)$''. This substitution may lead to training instability and exploding gradient issues. Recently, AI-Feynman~\cite{udrescu2020ai, udrescu2020ai2} was developed to decompose the process of finding an equation into a flow based on the assumption of known physical properties. However, this method relies heavily on prior physics knowledge, such as symmetries or invariances. A more recent approach, Deep Symbolic Regression (DSR)~\cite{petersen2019deep}, combined recurrent neural networks (RNN) with reinforcement learning for symbolic regression. Despite outperforming many GP-based approaches, DSR struggles with equations that contain multiple variables and constants.

\textbf{Tree-based Symbolic Regression.} Furthermore, a few recent studies proposed Monte Carlo tree search (MCTS)~\cite{cazenave2013monte,coulom2007efficient,lu2021incorporating,sun2022symbolic} for symbolic regression. The MCTS is performed in the following four steps: 1) selection, 2) expansion, 3) simulation, and 4) backpropagation. It takes advantage of the trade-off between exploration and exploitation to better discover mathematical expressions. For instance, a most recent work developed Symbolic Physics Learner (SPL)~\cite{sun2022symbolic} to accelerate discovery based on prior physics knowledge. However, SPL does not scale well to mathematical equations with many variables.

\textbf{Pretraining-based Symbolic Regression.} Inspired by large language models, researchers also adopted a pre-training technique based on Transformer~\cite{valipour2021symbolicgpt,lample2019deep,biggio2021neural} for the discovery of symbolic equations. For example, Biggio et al.~\cite{biggio2021neural} developed a large scale pre-training model for symbolic regression. To overcome the ill-posed problem in skeleton prediction, recent work developed an end-to-end (E2E) symbolic regression by training Transformer on a large amount of synthetic data. However, Transformer-based symbolic regression requires a ton of training data, which is not practical in real world applications. Moreover, it does not scale well to high-dimensional functions with many variables.

% This work will develop a simple and effective neural symbolic regression method using control variables to speed up discovery and improve estimation accuracy.

\section{Proposed Method}\label{sec:model}
In this section, we first state the problem of symbolic regression, and then elaborate on the proposed ScaleSR. A walk-through example is provided to enhance the understanding of our approach. Furthermore, we study how the proposed method effectively reduces the search space in symbolic regression.

%%%----problem statement----
\subsection{Problem Statement} 
% Given a set of $N$ data samples $\mathcal{D}=\{\mathbf{x}_i, y_i\}_{i=1}^N$, where $\mathbf{x}_i \in \mathbb{R}^d$ and $\mathbf{y} \in \mathbb{R}^{d_o}$. Here $d$ and $d_o$ denote the dimension of input and output data, respectively. The goal of symbolic regression is to learn an analytical mathematical expression, $\mathbf{y}=f(\mathbf{x}_i)$, \hongjue{$\bf{f}$ or $f$?} based on observed data $\mathcal{D}$. 

Given a set of $N$ data samples $\mathcal{D}=\{\mathbf{x}^{(n)}, y^{(n)}\}_{n=1}^N$, where $\mathbf{x}^{(n)} \in \mathbb{R}^d$ and ${y}^{(n)} \in \mathbb{R}$. Here $d$ denotes the dimension of input data. The goal of symbolic regression is to learn an analytical mathematical expression, $y=f(\mathbf{x})=f(x_1,x_2,\dots,x_d)$, based on observed data $\mathcal{D}$.

%%%-------overview----
\subsection{Proposed ScaleSR}
To improve the accuracy and scalability for multi-variable SR, we propose a scalable neural symbolic regression model to decompose it into a set of single-variable SR problems. The key idea is to learn a data generator from observed data using DNNs, and then use it to generate data samples by manipulating an independent variable each time. After that, we estimate the symbolic equation of the current variable based on its generated samples and then combine the discovered equations by adding variables one by one. Fig.~\ref{fig:framework} shows the overall framework of the proposed ScaleSR, which consists of three main parts: i) data generator with DNNs; ii) data generation via control variables; iii) single-variable symbolic regression (SR). Below, we will describe these three components in detail.

%%%---data generator---
\textbf{Data Generator with DNNs}. In many real-world applications, we only obtain the data samples from multiple input variables, rather than from a single control variable. In order to control data generation for a single variable, we first need to learn a data generator using deep neural networks (DNNs). After learning the mapping function between the input and output, $f(x_1,x_2,\dots,x_d)$, we can manipulate the input variables to generate different data samples as needed. For instance, we can vary variable $x_1$ while keeping the other variables fixed to generate data for $x_1$, i.e., $f_{x_2,\dots,x_d}(x_1)$. 

%% ----------part 2-------------
% \textbf{Data Generation via Control Variables.} For this part, we aim to generate different data samples by controlling the input variables. As mentioned earlier, our goal is to decompose symbolic regression (SR) for multiple variables into a subset of single-variable SR problems. To this end, we first use the above DNN generator to generate data points for the first\hongjue{? In the following parts, the article mentions that $x_1, \dots, x_{i-1}$ are already learned.} variable, e.g., $x_i$. Namely, we will generate a large number of data points $f_{x_1,x_2,\dots,x_N}(x_i)$\hongjue{$f_{x_1, x_2, \dots, x_{i-1}, x_{i+1}, \dots, x_N}(x_i)$} given different $x_i$. Then, we apply single-variable symbolic regression to estimate the corresponding mathematical expression. After that, as shown in Fig.~\ref{fig:data_gen}, we will add another independent variable $x_{i+1}$ to generate $M$ data samples, denoted by $f_M$, by randomly assigning different values to previously learned variables, i.e., $x_1, x_2, \dots,x_i$, while keeping the other variables $x_c$ fixed. By randomly choosing $K$ different values for newly added variable $x_{i+1}$, we can generate $K$ groups of data samples related to $x_{i+1}$. Here we use $\mathbf{F}^k$ to denote the $k$-th group of data points for the added variable $x_{i+1}$, and $\mathbf{X}^k$ to denote the remaining variables. Next, we will implement bottom-up symbolic regression to learn the mathematical equations for variable $x_{i+1}$. \hongjue{The notation $x_c$ may be a little confusing.}

\begin{figure*}[!tb]
\begin{center}
\includegraphics[width=\textwidth]{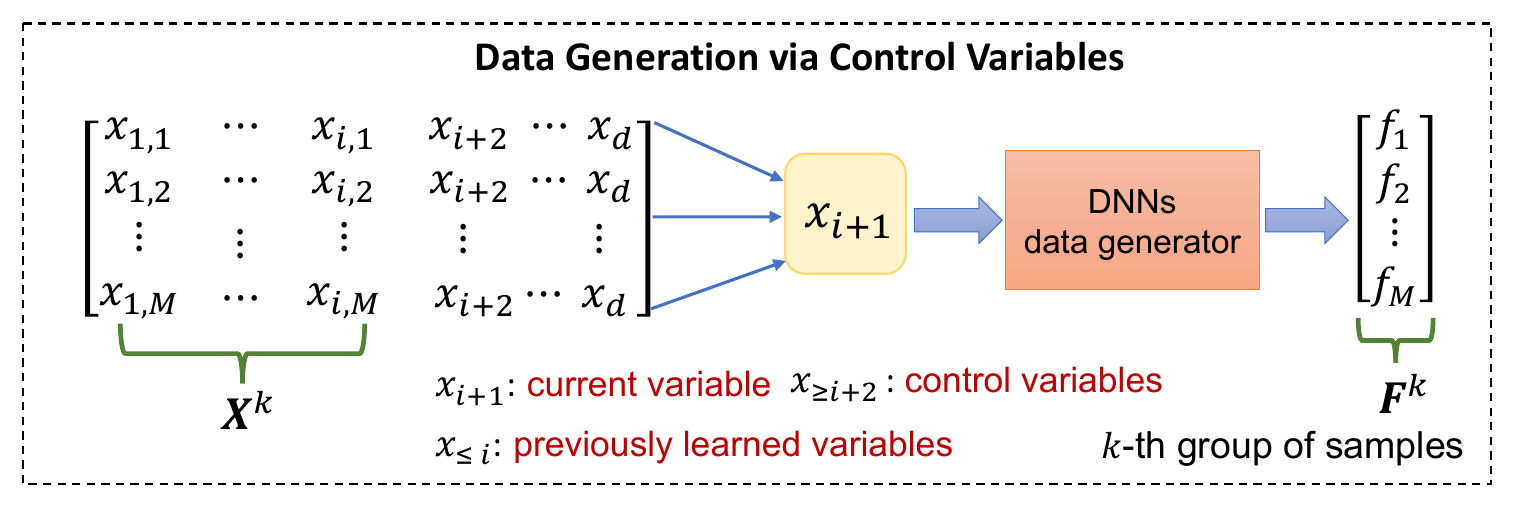}
 \vspace{-0.25in}
\caption{The framework of data generation with control variables. Specifically, we generate a group of data points for a newly added variable $x_{i+1}$ assigned with a random value and then vary the previously learned variables while fixing other control variables. By choosing $K$ different values for the current variable $x_{i+1}$, we can generate $K$ groups of data samples.}\label{fig:data_gen}
\vspace{-0.2in}
\end{center}
\end{figure*}

\textbf{Data Generation via Control Variables.} For this part, we aim to generate different data samples by controlling the input variables. As mentioned earlier, our goal is to decompose multi-variable SR into a set of single-variable SR problems. Suppose that we have learned a symbolic equation of the prior $i$ variables, denoted by $x_{\leq i}$ ($i=1,2,\dots)$. Next, we will estimate the mathematical equation of a newly added variable $x_{i+1}$. To achieve this, we use the above data generator to generate $K$ groups of data samples for the current variable $x_{i+1}$. For each group, we will generate $M$ data samples via varying the previously learned variables, i.e., $x_{\leq i}$, given a specific value of $x_{i+1}$, as shown in Fig.~\ref{fig:data_gen}. Specifically, we randomly assign $M$ different values to $x_{\leq i}$ while keeping other control variables $x_{\geq i+2}$ fixed and assigning a value to $x_{i+1}$. Then they will be fed into the data generator, denoted by $f_{x_{\geq i+2}}(x_{\leq i},x_{i+1})$, to produce $M$ samples for a given $x_{i+1}$. Here we use $\mathbf{F}^k$ to represent the $k$-th group of samples for $x_{i+1}$, and $\mathbf{X}^k$ to represent different values of previously learned variables $x_{\leq i}$. By randomly choosing $K$ different values for $x_{i+1}$, we can generate $K$ groups of data samples $\mathbf{F}=\{\mathbf{F}^k\}_{k=1}^K$. Our next step is to perform single-variable symbolic regression to estimate the expression of $x_{i+1}$ based on the generated samples.

%%-----overall framework------

\begin{figure*}[!tb]
\begin{center}
 \includegraphics[width=\textwidth]{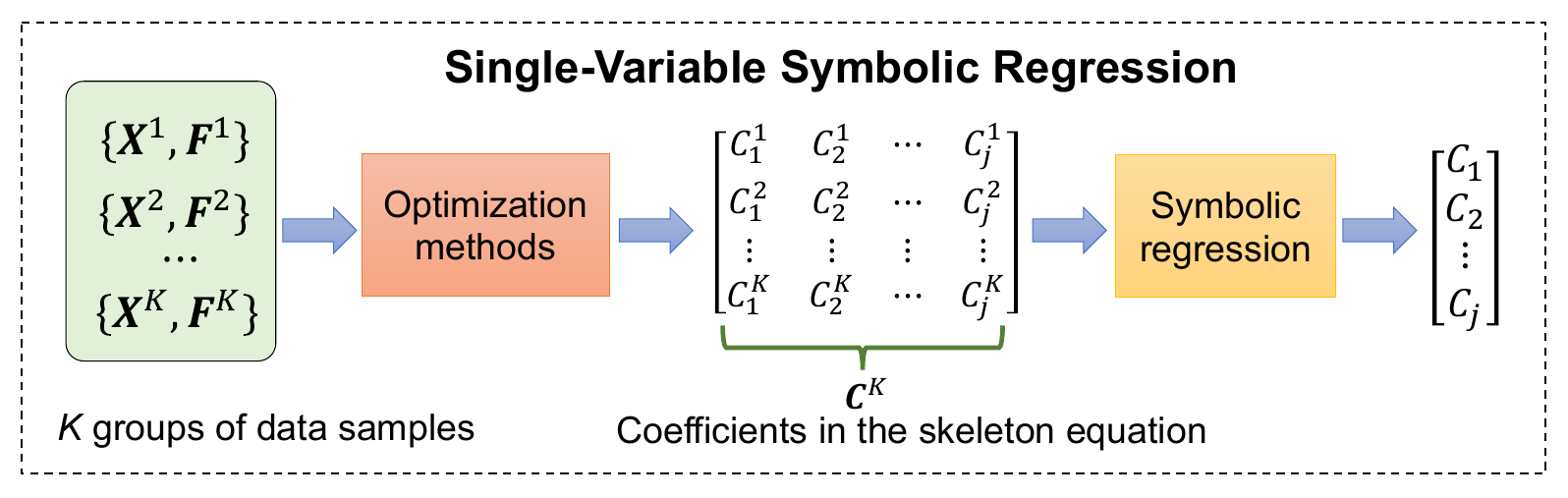}
 \vspace{-0.2in}
\caption{The framework of single-variable SR. It is performed in two steps: 1) we first use an optimization method, such as BFGS, to estimate $K$ groups of coefficients for the current independent variable; 2) we then use a single-variable SR model to estimate coefficients in the mathematical equation related to the current variable.}
\label{fig:bottom-up}
\vspace{-0.2in}
\end{center}
\end{figure*}

%%---------part 3-------------
\textbf{Single-Variable Symbolic Regression}. We propose single-variable SR to predict the mathematical expression for the current independent variable $x_{i+1}$. The key idea is to estimate the coefficients in the skeleton of previously learned variables, e.g., $f_{x\geq i+1}(x_{\leq i})=C_1x_i+C_2x_{i-1}x_1+\dots + C_j$, using the generated samples of $x_{i+1}$. As illustrated in Fig.~\ref{fig:bottom-up}, our approach is performed in two steps. (1) We adopt optimization techniques, such as BFGS, to estimate $K$ groups of coefficients $\mathbf{C}^K=\{C_1^k,\dots, C_j^k\}_{k=1}^K$ in the skeleton using $K$ groups of data samples $\{\mathbf{F}^k\}_{k=1}^K$ and the corresponding values of previously learned variables $\{\mathbf{X}^k\}_{k=1}^K$. Here, the coefficient $C_j$ in the skeleton can be viewed as a function of variable $x_{i+1}$. This step enables us to obtain $K$ groups of data samples $\mathbf{C}^K$ related to variable $x_{i+1}$ by manipulating it with $K$ different values. (2) We then apply symbolic regression to estimate the mathematical expression about $x_{i+1}$ given $K$ groups of $\mathbf{C}^K$ in the first step. Specifically, we feed the coefficient matrix $\mathbf{C}^K$ and the corresponding $K$ different values of $x_{i+1}$ into a symbolic model to estimate its skeleton and the corresponding coefficients, $\{C_1,\dots,C_j\}$. Finally, we repeat the above two steps by adding variables one by one until all the variables are covered.

%%%------

%%%-------walk through example-------
\subsection{A Walk-through Example}
To better understand the proposed method, we use a walk-through example to explain its core idea. Take $y= x_1x_2+2x_2+2$ as an example. Given a set of data points $\{x_1^{(n)},x_2^{(n)},y^{(n)}\}_{n=1}^N$, we first adopt DNNs to learn a mapping function $f(x_1,x_2)$ between the input variables $x_1$, $x_2$ and the output $y$, which will serve as a data generator. Then we use the data generator to generate different data samples for the independent variable $x_1$ by varying $x_1$ while keeping variable $x_2$ unchanged (e.g., $x_2=2$), i.e., $f_{x_2}(x_1)$. Next, we leverage a symbolic regression model, such as GP and MCTS, to estimate the mathematical equation about $x_1$, e.g., we get $f_{x_2}(x_1)= 2 x_1+6$. Since it is hard to directly derive $x_2$ from the discovered equation $f_{x_2}(x_1)$, we need to convert it into the following skeleton, $f_{x_2}(x_1)= C_1 x_1+C_2$, where $C_1$ and $C_2$ can be viewed as a function of $x_2$ that need to be estimated later. After that, we add another independent variable $x_2$ to the data generator $f(x_1,x_2)$, and then generate $M$ data samples given a random value of $x_2$, as shown in Fig.~\ref{fig:data_gen}. By choosing $K$ different values of $x_2$, we can generate $K$ groups of data samples, denoted by $\{\mathbf{F}^k(x_1,x_2)\}_{k=1}^K$. The next step is to use an optimization method, such as BFGS, to estimate the $k$-th group of coefficients $C_1^k$ and $C_2^k$ in the skeleton $f_{x_2}(x_1)$, given $\mathbf{F}^k(x_1,x_2)$ and $\mathbf{X}^k=[x_{1,1},x_{1,2},\dots,x_{1,M}]^\top$. Finally, we apply single-variable symbolic regression to estimate the symbolic regression about $x_2$ given $K$ groups of coefficients $\{C_1^k\}_{k=1}^K$ and $\{C_2^k\}_{k=1}^K$, as shown in Fig.~\ref{fig:bottom-up}. For instance, we can get $C_1=x_2$ and $C_2=2x_2+2$. Since there are no remaining variables, we complete the process of discovering symbolic equation. If there are additional variables, we repeat this process to estimate their symbolic expressions until all the variables are covered.

\subsection{Reduction of Search Space}
We also analyze the relationship between the complexity of a mathematical expression and search space, and then illustrate that the proposed method can significantly reduce the search space. In this work, the complexity of an expression is defined below.
\begin{definition}
Following prior work~\cite{petersendeep}, complexity is defined as twice the number of binary operators \{$+$, $-$, $\times$, $\div$\}, denoted by $N_b$, plus the number of unary operators \{$\sin$, $\cos$, $\exp$, $\log$\}, denoted by $N_u$, in the equation. Mathematically, the complexity can be formulated as $2N_b+N_u$.
\end{definition}
%%%%%%%%%%figure%%%%%%%%%%%%%%%
\begin{wrapfigure}[21]{r}{0.51\textwidth}
\vspace{-0.2in}
  \begin{center}
    \includegraphics[width=0.5\textwidth]{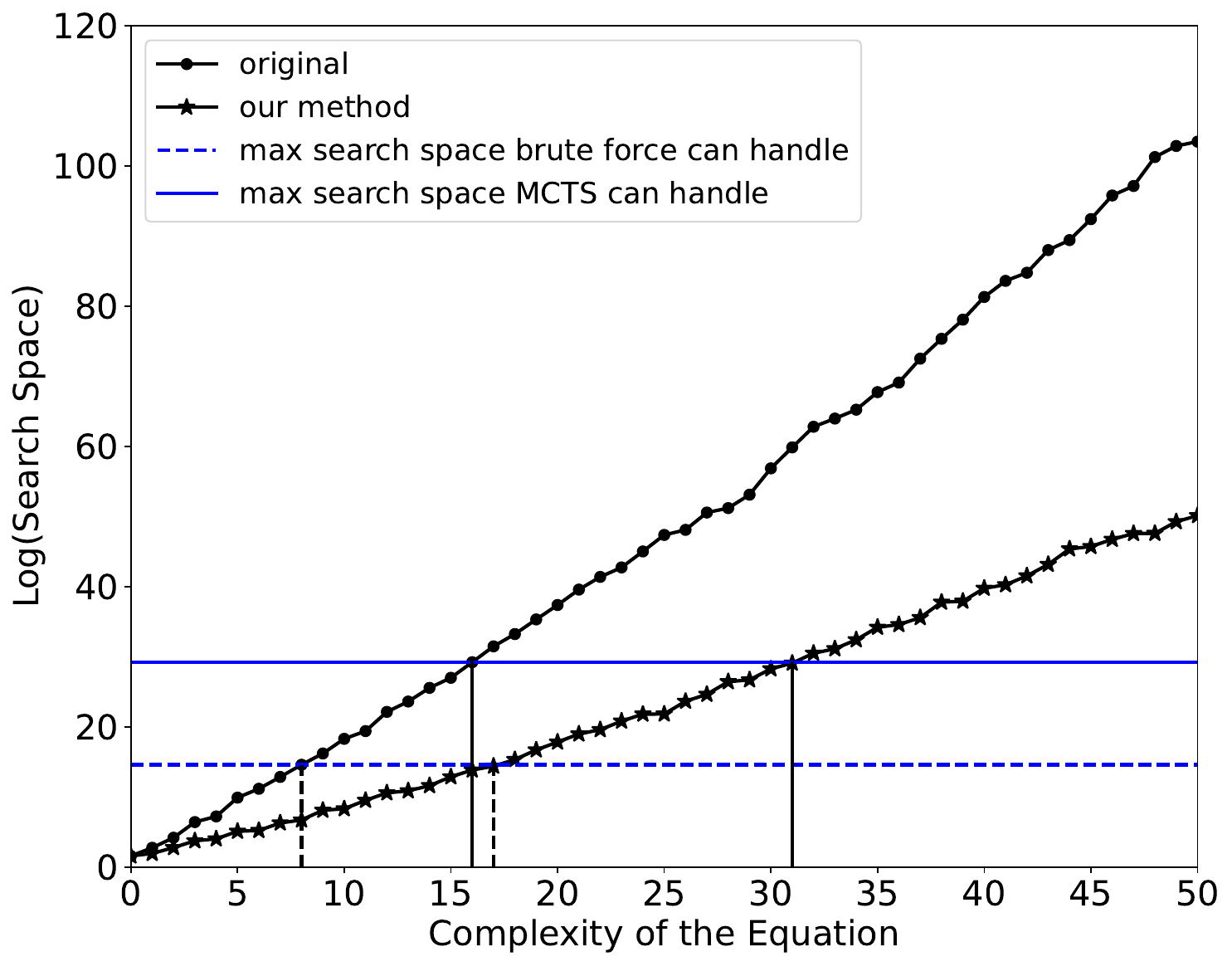}
  \end{center}
  \vspace{-0.15in}
\caption{The relationship between complexity and search space for different methods based on $1000$ equations with different complexity.}
\label{fig:compare_reduction}
\end{wrapfigure}

The main reason why we define the above complexity is that it is identical to the number of nodes in an expression tree minus one. Plus, most existing symbolic regression and brute force methods often adopt the expression tree for heuristic searching. Hence, we can use this metric to measure the difficulty of symbolic regression. 

% For instance, $\sin x_1+(C+x_1 \times x_1)$ has $3$ binary operators and $1$ unary operators, so its complexity is $7$.
%%%%%%%%

% \shao{under the same search space, our method can discover more complicated equations with multiple variables}
Fig.~\ref{fig:compare_reduction} shows the relationship between complexity and search space for our method and the state-of-the-art MCTS based on $1000$ equations with different complexity. Regarding how to sample these equations, please refer to the detailed description in Appendix~\ref{app:equations}. We can see from the two black curves search space will rise as the complexity is increased. Our method can significantly reduce the search space for discovering the same equation compared to the original MCTS in~\cite{sun2022symbolic}. The blue dashed line and solid line respectively represent the brute force and MCTS. We can see that our method can discover more complex equations under the same search space for both brute force and MCTS. For example, the original MCTS can discover an equation with a complexity of 16, while our method can estimate an equation with a complexity of 31, as shown in Fig.~\ref{fig:compare_reduction}.

\subsection{Algorithm Summary}
We summarize the proposed method in Algorithm~\ref{alg:sr}. We first learn a data generator from observed data using DNNs in Line 3. Lines 5-15 aim to generate $K$ groups of data samples $\mathbf{F}=\{\mathbf{F}^k\}_{k=1}^K$ and $\mathbf{X}=\{\mathbf{X}^k\}_{k=1}^K$ by manipulating the current variable $x_{i+1}$ with $K$ different values. Then, we use optimization methods to estimate the coefficients $\mathbf{C}^K$ in the skeleton based on $\mathbf{F}$ and $\mathbf{X}$ in Line 16. In Line 17, we apply single-variable SR to estimate the symbolic equation of $x_{i+1}$ based on $K$ groups of $\mathbf{C}^K$ and the current variable. We will repeat this process until all the variables are completed.
%% ----algorithm---
\begin{algorithm}[htb]\small
\caption{Proposed ScaleSR}\label{alg:sr}
\begin{algorithmic}[1]
\STATE \textbf{Input:} $N$ data samples $\mathcal{D}=\{\mathbf{x}^{(n)}, y^{(n)}\}_{n=1}^N$.
\STATE \textbf{Initial:} Previously learned variables $X_v=\{\}$ and controlled variables $X_c=\{x_{1},\dots,x_d\}$.
\STATE Learn a data generator using DNNs, $f(x_1,\dots,x_d)$, based on data $\mathcal{D}$;
\FOR{$i = 0, \dots, d-1$}
\STATE $X_c \gets X_c - \{x_{i+1}\}$ // remove one controlled variable;
\STATE /*Data generation via manipulating the newly added variable $x_{i+1}$*/
\STATE Assign random values to controlled variables in $X_c$;
\STATE $\mathbf{F}=\{\}$, $\mathbf{X}=\{\}$, $\mathbf{X}_{i+1}=\{\}$;
\FOR{$k = 1, \dots, K$}
\STATE Assign a random value to $x_{i+1}$ for each $k$;
\STATE Generate a group of samples $\mathbf{F}^k$ for the current variable $x_{i+1}$; Each group generates $M$ samples about previously learned variables in $X_v$, denoted by $\mathbf{X}^k$;
\STATE $\mathbf{F} \gets \mathbf{F} + \{\mathbf{F}^k\}$;
\STATE $\mathbf{X} \gets \mathbf{X} + \{\mathbf{X}^k\}$;
\STATE $\mathbf{X}_{i+1} \gets \mathbf{X}_{i+1} + \{x_{i+1}\}$;
\ENDFOR
\STATE Use optimization methods to estimate the coefficient matrix $\mathbf{C}^K$ based on $K$ groups of samples $\mathbf{F}$ and $\mathbf{X}$;
\STATE Apply single-variable SR to estimate the symbolic equation of $x_{i+1}$ based on $\mathbf{C}^K$ and $\mathbf{X}_{i+1}$;
\STATE $X_v \gets X_v + \{ x_{i+1} \}$  // add one variable;
\ENDFOR
\STATE \textbf{Output:} Discover analytical mathematical expression $y=f(\mathbf{x})$.
\end{algorithmic}
\end{algorithm}

% \chu{see last page}

% \begin{algorithm}[!tb]
%    \caption{PI algorithm.}
%    \label{alg:pid}
% \begin{algorithmic}[1]\small
%    \STATE {\bfseries Input:} desired KL $v_{kl}$, coefficients $K_p$, $K_i$, max/min value $\beta_{max}$, $\beta_{min}$, iterations $N$
%    \STATE {\bfseries Output:} hyperparameter $\beta(t)$ at training step $t$
% %   %\REPEAT
%    \STATE {\bfseries Initialization}: $I(0)=0$, $\beta(0)=0$
%    \FOR{$t=1$ {\bfseries to} $N$}
%    \STATE Sample KL-divergence, $\hat{v}_{kl}(t)$
%    \STATE $e(t) \leftarrow v_{kl}-\hat{v}_{kl}(t)$
%    \STATE $P(t) \leftarrow \frac{K_p}{1+\exp(e(t))}$
%    \IF{ $ \beta_{min} \leq \beta(t-1) \leq \beta_{max}$ }
%    		\STATE $I(t) \leftarrow I(t-1) - K_i e(t)$
%    \ELSE
%    \STATE $I(t) \leftarrow I(t-1)$ \quad // Anti-windup
% 	 \ENDIF
%    \STATE $\beta(t) \leftarrow P(t)+I(t) + \beta_{min}$
% %   \STATE $\beta(t-1) \leftarrow \beta(t)$
% %   \STATE $I(t-1) \leftarrow I(t)$
%    \IF{$\beta(t)> \beta_{max}$}
%    \STATE $\beta(t) \leftarrow \beta_{max}$
%  	\ENDIF
%    \IF{$\beta(t)< \beta_{min}$}
%    \STATE $\beta(t) \leftarrow \beta_{min}$
%    \ENDIF
%    \STATE \textbf{Return} $\beta(t)$
%  \ENDFOR
% \end{algorithmic}
% \end{algorithm}

\section{Experiment}\label{sec:experiment}
In this section, we carry out extensive experiments to evaluate the performance of ScaleSR. We first compare the discovery rate of our method with state-of-the-art baselines on two SR benchmarks. Next, we apply ScaleSR to identify the governing equations of two gene regulatory networks. Finally, we perform ablation studies to explore the impact of certain hyper-parameters on symbolic regression.

\subsection{Datasets}
We use two SR benchmark datasets, Nguyen~\cite{uy2011semantically} and Jin~\cite{jin2019bayesian}, for the first set of experiments. To illustrate the effectiveness of our method on complex regression, we specifically select equations containing at least two variables. We also evaluate our method on two gene regulatory networks, including the genetic toggle switch and the repressilator, using synthetic data. Detailed descriptions of these datasets are presented in Appendix~\ref{app:data}.

\subsection{Baselines}
Four baseline approaches are used for comparison with the proposed ScaleSR.
\begin{itemize}[leftmargin=2em,noitemsep,topsep=0pt]
\item Symbolic Physics Learner (SPL)~\cite{sun2022symbolic}. This method incorporates prior knowledge into Monte Carlo tree search for scientific discovery. 
\item Deep Symbolic Regression (DSR)~\cite{petersendeep}. It combines RL-based search method and recurrent neural networks (RNN) for symbolic regression. 
\item Gplearn (GP)~\cite{sipper2023ec}. It is a classic genetic programming method implemented in Python.
\item Neural-Guided Genetic Programming (NGGP)~\cite{mundhenk2021symbolic}. It is a hybrid method that combines RNN with GP for symbolic regression.
\end{itemize}

%%%%%%%%%%%%%%%%%%%%%%%%%%%%%%%%%%%%%%%%%%%%%
\subsection{Experimental Setup}
In the experiments, we have a pre-defined list of basic operations ($+, -, \times, \div, \text{const}$) and basic functions ($\sin, \cos, \exp, \log$). For SR benchmarks, we generate $N=8000$ data samples and then split them into $6400$ and $1600$ for training and validation, respectively. The proposed ScaleSR aims to discover the underlying mathematical expressions from data based on the above two lists of candidate operations. The discovered equations will be compared with the ground-truth expressions. For the data generator, 
we use three fully connected layers (MLP) with hidden sizes of 128, 256, and 128, respectively. Then we train the MLP using Adam optimizer with an initial learning rate of 0.1 and cosine annealing schedule. In addition, we use a single batch containing all input data due to the small number of training samples. For single-variable symbolic regression, we choose $M=200$ data samples for the current independent variable with $K=200$ different values. Also, we adopt MCTS in the prior work~\cite{sun2022symbolic} to estimate the symbolic equation with a single variable. This paper will use these hyperparameters in the following experiments, unless specified otherwise. Note that we will conduct ablation studies to investigate the impact of some important hyperparameters on the prediction performance of our method.

% Different test case uses different sets of candidate operations. There are four different candidate operation sets. The first one is the basic operation set $S_o = \{+, -, \times, \div, \text{const}\}$ which is used by all benchmarks. Additionally, there are three pairs of extra operation set $S_f = \{\bullet^2, \bullet^3\}$, $L_2 = \{\sin, \cos\}$, $L_3 = \{\log, \exp\}$. We provide SR models for these extra operation sets only when ground truth expressions contain these operations. ScaleSR and all baselines are fed by the same candidate operation set for the same test cases.

% Each test case is composed of a ground truth equation, a training data set, a testing data set, and a candidate operation set. The training data set and the candidate operation set are provided to the symbolic regression model to discover possible expressions. The ground truth equation and the testing data set are used to determine if the discovery equation is correct.

\noindent \textbf{Evaluation Metrics.}
We run $10$ independent tests for each case and calculate the recovery rate for each model. A successful discovery is evaluated using the following two criteria: i) prediction precision and ii) equation equivalence to ground truth. First, the mean square relative error (MSRE) between the prediction and ground truth should be less than $10^{-3}$. Second, the discovered equation should be in an identical or equivalent symbolic form to the target equation. We manually check the discovered symbolic equations to ensure their correctness.

% The criteria of a successful discovery concerning both the discovered equation's predicted precision and its symbolic equivalence to the ground-truth target. Only expressions with perfect fitness could be considered correct. To meet this requirement, we adopt the most strict metric that 1) the resulting equation should have a mean square relative error less than $10^{-4}$, and 2) it should have the exactly same symbolic form as the ground truth. To ensure the correctness, we manually check the discovered symbolic equations.

% Very few equations that pass the bound are in the wrong form, and the opposite, very few equations that have the right form fail the predicted error criteria. 

% \footnote{DSR and NGGP: \url{https://github.com/brendenpetersen/deep-symbolic-optimization} \\ SPL: \url{https://github.com/isds-neu/SymbolicPhysicsLearner}}.
\subsection{Evaluation on SR Benchmarks}
First, we evaluate the proposed ScaleSR on two widely used SR benchmarks: Nguyen and Jin. Table~\ref{tab:comparison_sr} illustrates the comparison of discovery rate for different methods using 10 random seeds. It can be observed that our method achieves higher recovery rates than the baselines. This is because the proposed ScaleSR adopts the similar idea of ``divide and conquer'' that decomposes multi-variable SR into a subset of single-variable SR problems. Besides, our method can significantly reduce the search space of symbolic regression, thus speeding up the discovery. Please refer to the comparison of computational cost in Appendix~\ref{app:cost}.

%%%%%%%%%%%%%%%%-----table--------%%%%%%%%%%%%%%%%%%%%%%%%
\begin{table}[!thb]
\centering
\caption{Recovery rate comparison of the proposed ScaleSR and other baselines on SR benchmarks, Nguyen and Jin. Our method significantly outperforms the baselines in terms of the average recovery rate. Note that GP does not work well since it only learns an approximated equation.}
\label{tab:comparison_sr}
\begin{adjustbox}{width=0.95\textwidth}
\begin{tabular}{lcccccc}
\hline
Benchmark & Expression & ScaleSR (ours) & SPL & NGGP & DSR & GP \\ \hline
Nguyen-09 & $\sin(x) + \sin(y^2)$ & 90\% & 100\% & 40\% & 30\% & 20\% \\
Nguyen-10 & $2\sin(x)\cos(y)$ & 90\% & 70\% & 100\% & 100\% & 90\% \\
Nguyen-11 & $x^y$ & 70\% & 70\% & 100\% & 90\% & 0\% \\
Nguyen-12 & $x^4 - x^3 + \frac{1}{2}y^2 - y$ & 70\% & 30\% & 10\% & 0\% & 10\% \\ \hline
Jin-1 & $2.5 x^4 - 1.3 x^3 + 0.5 y^2 - 1.7 y$ & 70\% & 0\% & 0\% & 0\% & 0\% \\
Jin-2 & $8.0 x^2 + 8.0 y^3 - 15$ & 100\% & 90\% & 70\% & 50\% & 10\% \\
Jin-3 & $0.2 x^3 + 1.5 y^3 - 1.2 y - 0.5 x$ & 90\% & 90\% & 0\% & 0\% & 0\% \\
Jin-4 & $1.5 \exp(x) + 0.5 \cos(y)$ & 100\% & 100\% & 40\% & 10\% & 0\% \\
Jin-5 & $6.0 \sin(x) \cos(y)$ & 100\% & 80\% & 100\% & 100\% & 0\% \\
Jin-6 & $1.35 xy + 5.5 \sin((x-1.0)(y-1.0))$ & 0\% & 0\% & 0\% & 0\% & 0\% \\ \hline
Average &  & \textbf{78\%} & 63\% & 46\% & 38\% & 13\% \\ \hline
\end{tabular}
\end{adjustbox}
\end{table}

\subsection{Evaluation on Gene Regulatory Networks}
Next, we apply the proposed method to discover the underlying governing equations of two classic gene regulatory networks, the genetic toggle switch and the repressilator.

% \begin{wrapfigure}[23]{r}{0.51\textwidth}
%   \begin{center}
%     \includegraphics[width=0.5\textwidth]{figs/Toggle.pdf}
%   \end{center}
% \caption{}
% \label{fig:toggle}
% \end{wrapfigure}

%%%%%------model 1---%%%%%%%%%%%%%%
\textbf{Genetic toggle switch}. The genetic toggle switch \cite{gardner2000construction} is a synthetic gene regulatory network that has been extensively studied as a fundamental concept in the field of synthetic biology. It has numerous prospective applications in biotechnology, such as the development of biosensors, gene therapies, and synthetic memory devices. The genetic toggle switch consists of two mutually repressive genes controlled by their respective promoters, creating a bistable system that can be toggled between two stable states as follows.
% Understanding the principles of the toggle switch model can aid in the design of more complex gene regulatory networks used to engineer biological systems with specific tasks.
%%%%%%%%%%%%
\begin{equation}\small
\begin{cases}
\dfrac{dU}{dt} = \dfrac{\alpha_{1}}{1+V^{\beta}}-U\vspace{1ex} \\
\dfrac{dV}{dt} = \dfrac{\alpha_{2}}{1+U^{\gamma}}-V\vspace{1ex}, \\
\end{cases}
\label{eq:toggle}
\end{equation}
where $\alpha_{1}$ and $\alpha_{2}$ are the synthesis rates of repressors $U$ and $V$, respectively. $\beta$ and $\gamma$ are the cooperativities of repression on two promoters.

In this experiment, following the bistable region in prior work~\cite{gardner2000construction}, we choose $\alpha_{1}=4$, $\alpha_{2} = 4$, $\beta = 3$, $\gamma = 3$, and the initial conditions $U(0), V(0) \in [0, 4]$. To train our model, we generate $1000$ trajectories by randomly choosing $1000$ initial conditions. We use $800$ of them as training data and the remaining $200$ as validation data. The time span of each trajectory is $t \in [0, 1]$ with a sampling time interval of $0.01$. Namely, we sample 100 data points for each trajectory.

\begin{figure*}[!t]
\centering
\vspace{-0.15in}
\subfigure[Genetic toggle switch]{
\includegraphics[height=0.29\linewidth]{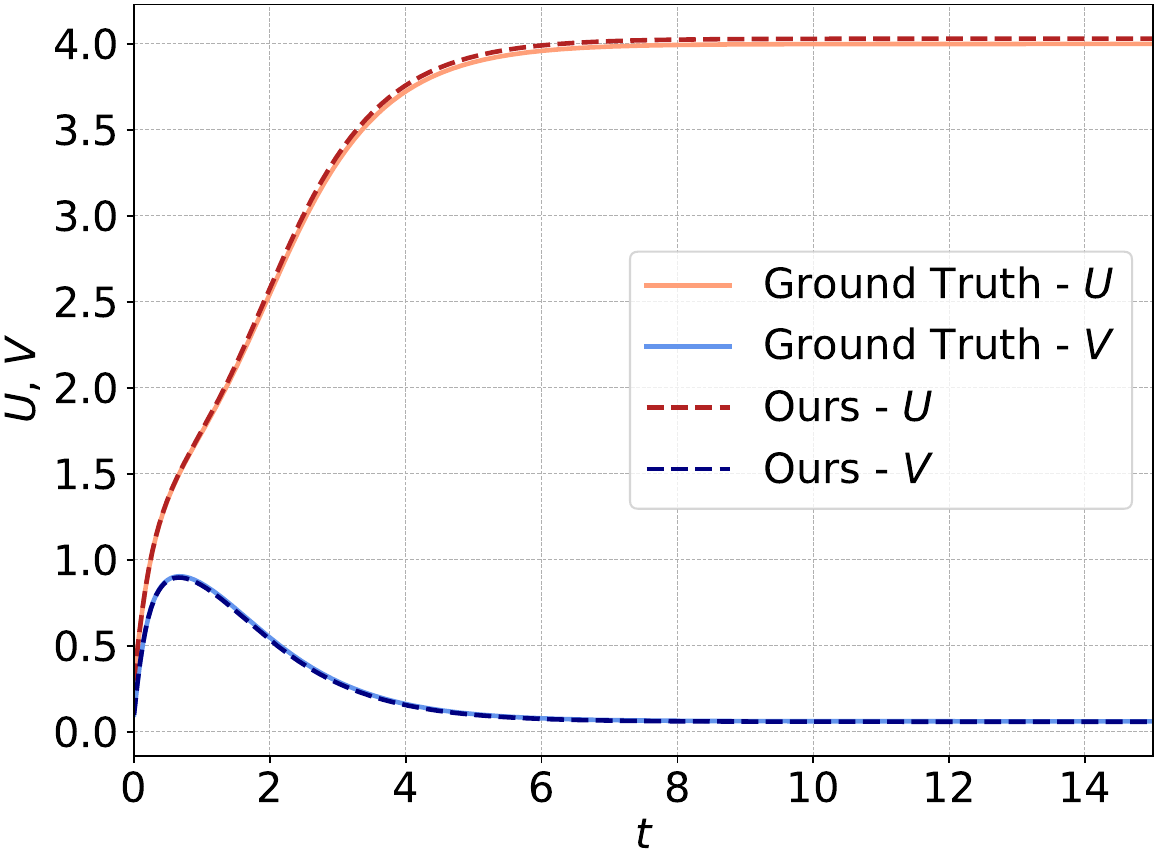}}
\subfigure[Repressilator]{
\includegraphics[height=0.29\linewidth]{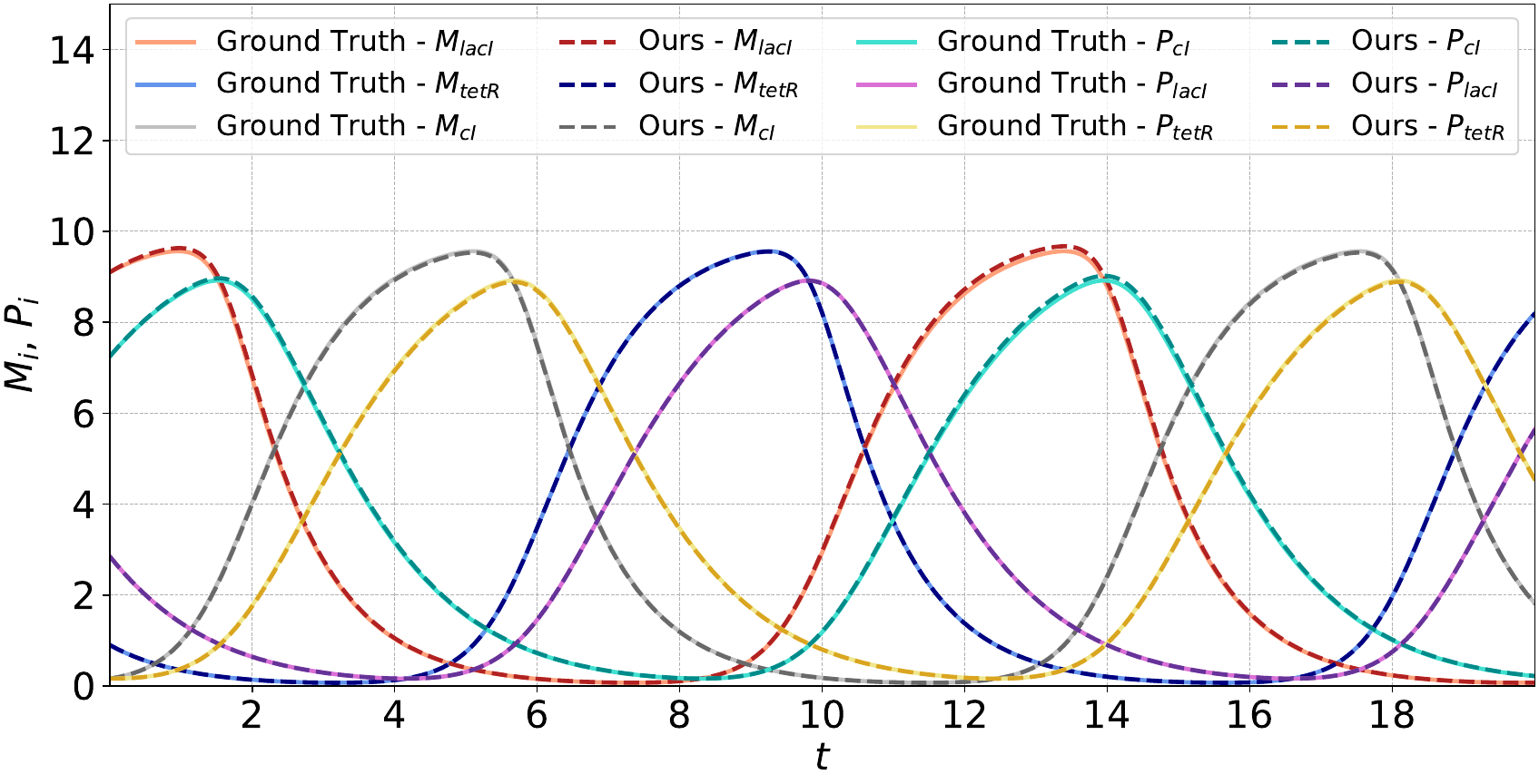}}
\vspace{-0.1in}
\caption{Trajectory prediction of the genetic toggle switch and the repressilator using ScaleSR. The ScaleSR can precisely predict their trajectories, closely matching the ground truth.}
\label{fig:bio_sys}
\vspace{-0.1in}
\end{figure*}

Fig.~\ref{fig:bio_sys} (a) illustrates the predicted trajectories of the genetic toggle switch using ScaleSR with a random initial condition. It can be observed that ScaleSR precisely predicts the trajectory, closely matching the ground truth obtained from the ODE solver (odeint). The mean square relative error (MSRE) of our method is about $9.03 \times 10^{-4}$. Importantly, our method successfully discovers the underlying governing equation from observed data as follows. We can see that it is quite close to the target model in Eqs.~\ref{eq:toggle}. We also present the experimental results of the \textit{baselines} in Appendix~\ref{app:baseline_bio}.
%%%%%%%%%%%%%%%%%%
\begin{equation}\small
\begin{cases}
\dfrac{dU}{dt} = \dfrac{3.919}{0.972+V^{3}}-U\vspace{1ex} \\
\dfrac{dV}{dt} = \dfrac{3.921}{0.972+U^{3}}-V\vspace{1ex}, \\
\end{cases}
\label{eq:toggle-my}
\end{equation}
%%%%%%%%%%%%%%%%%%

%%%%%------model 2---%%%%%%%%%%%%%%
\noindent
\textbf{Repressilator}.
The Repressilator \cite{elowitz2000synthetic} is another type of gene regulatory network that exhibits oscillatory behavior.
This model is critical in studying the dynamics of genetic circuits and provides insights into the principles of oscillatory systems in biology. Comprising three genes, it operates through a feedback loop in which each gene produces a repressor protein that suppresses the expression of the subsequent gene. This process results in a cyclic pattern of gene expression, described as follows.
\begin{equation}
\begin{cases}
\dfrac{dM_{i}}{dt} = - M_{i}+\dfrac{\alpha}{1+P^{n}_{j}}+\alpha_{0}\vspace{1ex} \\
\dfrac{dP_{i}}{dt} = - \beta\left(P_{i}-M_{i}\right)\vspace{1ex} 
\end{cases}
\begin{pmatrix} i=lacI, tetR, cI \\ j=cI, lacI, tetR \end{pmatrix},
\label{eq:repressilator6}
\end{equation}
In Eqs.~\ref{eq:repressilator6}, $P_{i}$ denotes the repressor protein concentrations, and $M_{i}$ represents the corresponding mRNA concentrations, where $i$ is $lacI$, $tetR$, or $cI$. If there are saturating amounts of repressor, the number of protein copies produced from a given promoter is $\alpha_{0}$. Otherwise, this number is $\alpha+\alpha_{0}$. $\beta$ represents the ratio of the protein decay rate to the mRNA decay rate, and $n$ is a Hill coefficient.

In this experiment, we set $\beta = 1$, $\alpha_0 = 10^{-5}$, $\alpha = 10$, $n = 3$, and the initial conditions $M_i, P_j \in [0, 5]$. To train our model, we generate $5000$ trajectories using $5000$ random initial conditions. They are split into $4000$ training data and $1000$ validation data respectively. The time span of each trajectory is $t \in [0, 4]$ with a sampling time interval of $0.01$.

Fig.~\ref{fig:bio_sys} (b) illustrates the trajectory predictions of repressilator using our method. We can see that the trajectory predicted by ScaleSR closely aligns with the ground truth, with MSRE of about $7.54\times 10^{-5}$. Additionally, our method successfully identifies the underlying governing equations, which are provided in Appendix~\ref{app:protein_equ}.

% figs/Protein.pdf
% \begin{wrapfigure}[23]{r}{0.54\textwidth}
%   \begin{center}
%     \includegraphics[width=0.5\textwidth]{figs/Protein.pdf}
%   \end{center}
% \caption{}
% \label{fig:protein}
% \end{wrapfigure}

% \noindent \textbf{Implementation details.}
% In this work, the proposed method is implemented in PyTorch~\cite{paszke2017automatic} while the baselines are implemented in PyCIL~\cite{zhou2021pycil}, which is a well-known toolbox for CIL. In our experiments, we trained ResNet-18~\cite{he2016deep} from scratch with the SGD~\cite{ruder2016overview} optimizer with an initial learning rate of 0.01. Then the learning rate is multiplied by 0.1 per 15 epochs. We train the model for 60 epochs with batch size 256. The experimental results are averaged over three random seeds. We execute different incremental phases (i.e., 5, 10 incremental phases) under both zero-base and half-base settings for CIFAR-100 and Tiny-ImageNet. As for the memory size of exemplar-based approaches: iCaRL~\cite{rebuffi2017icarl}, BiC~\cite{wu2019large}, PODNet~\cite{douillard2020podnet}, WA~\cite{zhao2020maintaining}, and FOSTER~\cite{wang2022foster}, we use \textit{herd election}~\cite{rebuffi2017icarl} to select exemplars of previous tasks in zero-base setting. For the half-base setting, we store up to $K = 2000$ exemplars the same setting as~\cite{rebuffi2017icarl} and we store 20 samples per old class following the common setting in ~\cite{hou2019learning}.

\subsection{Ablation Studies}\label{sec:ablation}
% We also conduct ablation studies to explore the impact of some hyper-parameters on the prediction accuracy of our method.

\noindent
\textbf{Effect of $M$ in data generation.}
First, we examine the impact of $M$ different values of previously learned variables in data generation on the recovery rate. As shown in Table~\ref{tab:ablation_M}, when $M$ varies from $50$ to $200$, the recovery rates of our approach remain fairly consistent. This suggests that our method is not sensitive to the number of generated samples $M$ as it is sufficiently large.

%%%%%%%%%%%%%%Ablation study%%%%%%%%%%%%
\begin{table}[htb]
\centering
\caption{Effect of $M$ samples on recovery rate of the proposed ScaleSR using Nguyen.}
\label{tab:ablation_M}
% \begin{adjustbox}{width=0.95\textwidth}
\begin{tabular}{lccccc}
\hline
Benchmark & Expression & M=50 & M=100 & M=150 & M=200  \\ \hline
Nguyen-09 & $\sin(x) + \sin(y^2)$ & 90\% & 90\% & 90\% & 90\% \\
Nguyen-10 & $2\sin(x)\cos(y)$ & 90\% & 90\% & 90\% & 90\%  \\
Nguyen-11 & $x^y$ & 70\% & 70\% & 70\% & 70\%  \\
Nguyen-12 & $x^4 - x^3 + \frac{1}{2}y^2 - y$ & 60\% & 70\% & 70\% & 70\% \\ \hline
\end{tabular}
% \end{adjustbox}
\end{table}

%%%-----train samples------------
\noindent
\textbf{Effect of $K$ groups in data generation.}
Next, we study the effect of $K$ groups of generated data samples for the current variable on the recovery rate of ScaleSR. We can observe from Table~\ref{tab:ablation_K} that the recovery rates almost keep the same as $K$ changes from $50$ to $200$. 

%%%%%%%%%%%%%%Ablation study%%%%%%%%%%%%
\begin{table}[htb]
\centering
\caption{Effect of $K$ groups on the recovery rate of our ScaleSR using Nguyen.}
\label{tab:ablation_K}
% \begin{adjustbox}{width=0.95\textwidth}
\begin{tabular}{lccccc}
\hline
Benchmark & Expression & K=50 & K=100 & K=150 & K=200  \\ \hline
Nguyen-09 & $\sin(x) + \sin(y^2)$ & 90\% & 90\% & 90\% & 90\% \\
Nguyen-10 & $2\sin(x)\cos(y)$ & 90\% & 90\% & 90\% & 90\%  \\
Nguyen-11 & $x^y$ & 70\% & 70\% & 60\% & 70\%  \\
Nguyen-12 & $x^4 - x^3 + \frac{1}{2}y^2 - y$ & 60\% & 70\% & 70\% & 70\% \\ \hline
\end{tabular}
% \end{adjustbox}
\end{table}

%%%%%
\noindent
\textbf{Effect of $N$ training samples.}
Lastly, we investigate the influence of the training data size ($N$) on the proposed ScaleSR. It can be seen that our method achieves good performance when the number of training samples is sufficiently large. If the equations are more complex, it might be necessary to increase the amount of training data provided to the DNNs for optimal results.

%%%%%%%%%%%%%%Ablation study%%%%%%%%%%%%
\begin{table}[!htb]
\centering
\caption{Effect of $N$ training samples on the recovery rate of our ScaleSR using Nguyen.}
\label{tab:ablation_N}
% \begin{adjustbox}{width=0.95\textwidth}
\begin{tabular}{lcccccc}
\hline
Benchmark & Expression & N=800 & N=1600 & N=3200 & N=4800 & N=6400 \\ \hline
Nguyen-09 & $\sin(x) + \sin(y^2)$ & 90\% & 90\% & 90\% & 90\% & 90\% \\
Nguyen-10 & $2\sin(x)\cos(y)$ & 90\% & 90\% & 90\% & 90\% & 90\% \\
Nguyen-11 & $x^y$ & 70\% & 70\% & 70\% & 70\% & 70\% \\
Nguyen-12 & $x^4 - x^3 + \frac{1}{2}y^2 - y$ & 60\% & 60\% & 60\% & 70\% & 70\% \\ \hline
\end{tabular}
% \end{adjustbox}
\end{table}

\section{Conclusion}
In this work, we developed two non-exemplar-based methods, YONO and YONO+, for class-incremental learning. Specifically, YONO only needs to store and replay one prototype for each class without generating synthetic data from stored prototypes. As an extension of YONO, YONO+ proposed to create synthetic replay data from stored prototypes via a high-dimensional rotation matrix and Gaussian noise. The evaluation results on multiple benchmarks demonstrated that both YONO and YONO+ can significantly outperform the baselines in terms of accuracy and average forgetting. In particular, the proposed YONO achieved comparable performance to YONO+ with data synthesis. Importantly, this work offered a new perspective of optimizing class prototypes for exemplar-free incremental learning.

%%%%%--------reference-------------------
\bibliographystyle{abbrv}
\bibliography{reference.bib}

%%%%%%%%%%%%%%%%%%%%%%%%%%%%%%%%%%%%%%%%%%%%%%%%%%%%%%%%%%%%
\clearpage
\appendix
% \section{Proof of Complexity in an Expression Tree}\label{app:complexity}
% model implement
\section{Generate Expression Tree in Fig.~\ref{fig:compare_reduction}}\label{app:equations}

We introduce how to generate a mathematical expression with a specific complexity to estimate the search space in symbolic regression. In this work, we attempt to sample equations uniformly, i.e., all \textit{valid} equations should have the same probability to be selected. A valid equation defined as follows.

\begin{definition}
An equation is \textit{valid} if it is a proper mathematical equation consisting of $M_t~(M_t = 5)$ terminal symbols $\{x_1, x_2, x_3, x_4, const\}$, $M_u~(M_u = 4)$ unary operators $\{\sin, \cos, \log, \exp\}$, and $M_b~(M_b = 4)$ binary operators $\{+, -, \times, \div\}$. Moreover, it should not contain nested unary operators, such as $sin(cos(x_1)+1)$  or $log(sin(x_1))$, since they are not very meaningful in most real cases.
\end{definition}

Before describing our sampling method, we first define $F_i$ and $G_i$ that will be used for sampling.

\begin{definition}\label{def:FG}
Let $F_i$ and $G_i$ be the number of valid equations with complexity $i$ containing no unary operator and at least one unary operator, respectively. The criterion for distinguishing two equations is based on the structure of their expression trees rather than their algebraic equivalence. According to this, some equations that are algebraically equivalent, such as $x_1+x_2+x_3$, will be counted multiple times due to different tree structures. Nevertheless, it almost has no influence on the overall results.
\end{definition}

Next, we introduce the idea of calculating $F_i$ and $G_i$ using dynamic programming. First, we consider $F_i$. When the complexity $i$ is 0, the valid expression set contains only $M_t$ terminals without unary operators, so $F_0 = M_t$. When the complexity is greater than or equal to 1, we need to consider the root of the expression tree. Since $F_i$ counts an expression tree with no unary operator, its root has to be a binary operator with $M_b$ possibilities. Suppose the left subtree has complexity of $j$ ($j=0, \dots, i - 2$), then the complexity of the right subtree is $i-j-2$. As a result, it has $F_j F_{i-j-2}$ feasible options given a certain $j$. Finally, we can sum up all the cases and multiply the result by $M_b$ to get $F_i$.

% Consider the root has $M_b$ choices in two scenarios (i) and (ii), we can 

% Consider the root has $M_b$ choices,  we sum up all the options for the above two , and then multiply the result by $M_b$, consider the root has $M_b$ choices.

\begin{equation}
F_i = 
\begin{cases}
M_t & i = 0 \\
M_b \sum_{j=0}^{i-2} F_j F_{i-j-2} & i > 0,
\end{cases} 
\label{eq:calc-f}
\end{equation}

For $G_i$, when the complexity is $0$, there is no unary operator, so $G_0 = 0$. When the complexity is greater than or equal to $1$, the root can be either a unary operator or a binary operator. (1) If the root is a unary operator, we have $M_u F_{i-1}$ options for tree structures. (2) If the root is a binary operator, we assume that the left subtree has complexity of $j$ ($j=0, \dots, i - 2$), then the right subtree has the complexity of $i-j-2$. According to the Definition~\ref{def:FG}, one or both of the subtrees should contain at least one unary operator, so the number of options for tree structures should be $ G(i,j) = G_j G_{i-j-2} + G_j F_{i-j-2} + F_j G_{i-j-2}$. Since the root has $M_b$ choices in this scenario, we can multiply $G(i,j)$ by $M_b$ to get the total options for a binary operator. Combining (1) and (2), we can get $G_i$ as follows.

\begin{equation}
G_i = 
\begin{cases}
0 & i = 0 \\
M_u F_{i-1} + M_b \sum_{j=0}^{i-2} G_j G_{i-j-2} + G_j F_{i-j-2} + F_j G_{i-j-2} & i > 0,
\end{cases}
\label{eq:calc-g}
\end{equation}

Lastly, we summarize the sample subroutine in Algorithm~\ref{alg:sample}. First, we need to figure out whether a mathematical expression contains a unary operator or not. The probability of containing at least one unary operator is $G_i / (F_i+G_i)$, and that of not containing one is $F_i / (F_i+G_i)$. If there is no unary operator, we use \textsc{SampleByNU} in Lines 5-17 to generate the expression. This function operates as follows: if the complexity is $0$, we select a terminal symbol $M_u~(M_u = 4)$ as mentioned above. Otherwise, we sample the operator at the root and the complexity of the left subtree according to the number of their options. After this, we recursively invoke subroutines to generate the subtrees. If there exists at least one unary operator, we use \textsc{SampleByU} instead, which uses a similar way to that in \textsc{SampleByNU}. By doing this, we can generate an expression tree with a specified complexity.

\begin{algorithm}[!htb]
\caption{Expression Tree Sampling Algorithm}\label{alg:sample}\small
\begin{algorithmic}[1]
\STATE \textbf{Input:} Target complexity $N$, and pre-calculated $F_i$, $G_i$
\FUNCTION{SampleByWeight}{$weights$}
    \STATE \textbf{return} $i$ with possibility $\frac{weights_i}{\sum weights}$. $weights$ indexed from 0.
\ENDFUNCTION
\FUNCTION{SampleNU}{$N$}
    \STATE // Sample equation with no unary operator according to Equation~\ref{eq:calc-f}
    \IF{$N = 0$}
        \STATE \textbf{return} one terminal symbol sampled from \{`$x1$', `$x2$', `$x3$', `$x4$', `const'\}  
    \ENDIF
    \STATE $weights$ $\leftarrow$ $\{\}$
    \FOR{$i = 0, \dots, N-2$}
        \STATE $weights$ $\leftarrow$ $weights$ $+$ $\{F_i \cdot F_{N-i-2}\}$ 
    \ENDFOR
    \STATE Sample $i$ using \textsc{SampleByWeight}($weights$)
    \STATE Sample $operator$ from \{`$+$', `$-$', `$\times$', `$\div$'\}
    \STATE \textbf{return} \textsc{SampleNU}($i$) $+$ $operator$ $+$  \textsc{SampleNU}($N-i-2$)
\ENDFUNCTION
\FUNCTION{SampleU}{$N$}
    \STATE // Sample equation with at least one unary operator according to Equation~\ref{eq:calc-g}
    \IF{\textsc{SampleByWeight}$(\{M_u F_{N-1}, G_N - M_u F_{N-1}\})$ $= 0$}
        \STATE Sample $operator$ from \{`$\sin$', `$\cos$', `$\log$', `$\exp$'\}
        \STATE \textbf{return} $operator$ $+$ \textsc{SampleNU}($N-1$)
    \ELSE
        \STATE Use the same method as \textsc{SampleNU} to sample a tree with binary operator as the root
    \ENDIF
\ENDFUNCTION
\IF{\textsc{SampleByWeight}$(\{F_N, G_N\})$ $= 0$}
\STATE Sample equation $Equ$ using \textsc{SampleNU}($N$)
\ELSE
\STATE Sample equation $Equ$ using \textsc{SampleU}($N$)
\ENDIF
\STATE \textbf{Output:} The sampled equation $Equ$
\end{algorithmic}
\end{algorithm}

\section{Dataset Description}\label{app:data}
First, we introduce how to generate training data and validation data using two SR benchmarks. As shown in Table~\ref{tab:dataset}, we choose a certain range and then generate $8000$ data samples. Then we split them into $6400$ and $1600$ for training and validation, respectively.

\begin{table}[!thb]
\centering
\caption{Detailed description of the SR benchmark datasets.}
\label{tab:dataset}
\begin{adjustbox}{width=0.8\textwidth}
\begin{tabular}{lccc}
\hline
Benchmark & Expression & Range & Num. of samples  \\ \hline
Nguyen-09 & $\sin(x) + \sin(y^2)$ & [-3, 3] & 8000 \\
Nguyen-10 & $2\sin(x)\cos(y)$ & [-3, 3] & 8000  \\
Nguyen-11 & $x^y$ & [1, 2] & 8000 \\
Nguyen-12 & $x^4 - x^3 + \frac{1}{2}y^2 - y$ & [-3, 3] & 8000 \\ \hline
Jin-1 & $2.5 x^4 - 1.3 x^3 + 0.5 y^2 - 1.7 y$ & [-3, 3] & 8000 \\
Jin-2 & $8.0 x^2 + 8.0 y^3 - 15$ & [-3, 3] & 8000 \\
Jin-3 & $0.2 x^3 + 1.5 y^3 - 1.2 y - 0.5 x$ & [-3, 3] & 8000 \\
Jin-4 & $1.5 \exp(x) + 0.5 \cos(y)$ & [-3, 3] & 8000 \\
Jin-5 & $6.0 \sin(x) \cos(y)$ & [-3, 3] & 8000 \\
Jin-6 & $1.35 xy + 5.5 \sin((x-1.0)(y-1.0))$ & [-3, 3] & 8000 \\ \hline
% Average &  & \textbf{78\%} & 63\%   \\ \hline
\end{tabular}
\end{adjustbox}
\end{table}

Next, we describe how to generate the synthetic data for two gene regulatory networks: the genetic toggle switch and
the repressilator. For the genetic toggle switch, we generate $1000$ trajectories by randomly choosing $1000$ initial conditions. We use $800$ of them as training data and the remaining $200$ as validation data. The time span of each trajectory is $t \in [0, 1]$ with a sampling time interval of $0.01$. Namely, we sample 100 data points for each trajectory. For the repressilator,  we generate $5000$ trajectories using $5000$ random initial conditions. They are split into $4000$ training data and $1000$ validation data respectively. The time span of each trajectory is $t \in [0, 4]$ with a sampling time interval of $0.01$.

\section{Experimental Setup for Different Methods}
In this subsection, we present the experimental settings for the baselines. Following the prior work~\cite{sun2022symbolic}, for gplearn, we set the population to be $10000$ and the number of generations to $50$. For other baselines, SPL, NGGP, and DSR, we directly use their source code with the default parameters to implement experiments.

%%%%%%%--computational cost-----------
\section{Computational Cost of Different Methods}\label{app:cost}
We also compare the computational cost of the proposed ScaleSR and baseline approaches on Nguyen benchmark, as shown in Table~\ref{tab:running-time}. Our experimental results illustrate that our method, including DNNs and single-variable SR, have less running time than the baseline on Nguyen-12. However, it needs more running time than GP on Nguyen-09 and Nguyen-10, than DSR and NGGP on Nguyen-10 and Nguyen-11. However, our method has much higher recovery rate than DSR and NGGP, as illustrated in Table~\ref{tab:comparison_sr}.

% It is also worth noting that our method can replace the MCTS for single-variable SR with the other symbolic regressions in order to speed up discovery.
% Importantly, the proposed ScaleSR can insert any symbolic regression models to estimate the expression of a single variable.

\begin{table}[!thb]
\centering
\caption{Comparison of running time (seconds)}
\label{tab:running-time}
\begin{adjustbox}{width=0.95\textwidth}
\begin{tabular}{lcccccc}
\hline
Benchmark & Expression & ScaleSR (ours) & SPL & NGGP & DSR & GP \\ \hline
Nguyen-09 & $\sin(x) + \sin(y^2)$ & 697.00 & 2773.32 & 2551.89 & 1080.58 & 36.01 \\
Nguyen-10 & $2\sin(x)\cos(y)$ & 675.46 & 2447.58 & 383.27 & 89.40 & 10.46 \\
Nguyen-11 & $x^y$ & 655.60 & 4674.99 & 131.48 & 45.32 & 999.91 \\
Nguyen-12 & $x^4 - x^3 + \frac{1}{2}y^2 - y$ & 820.44 & 4932.86 & 3428.79 & 3156.63 & 1754.50 \\ \hline
\end{tabular}
\end{adjustbox}
\end{table}

\section{Discovered Governing Equations of Repressilator}\label{app:protein_equ}
Below, we present the discovered equations of the repressilator using our method. We can see that the equations are very close to the target model, except for $\alpha_0 = 10^{-5}$. The main reason is that $\alpha_0$ is too tiny to be estimated. However, it does not impact the trajectory prediction too much, according to our results in Fig.~\ref{fig:bio_sys}. 
%%%%%%%%%%%%%%%%%%%%
\begin{equation}
\begin{cases}
\dfrac{dM_{lacI}}{dt} = -M_{lacI} + \dfrac{9.939}{0.982+P_{tetR}^{3}}\vspace{1ex} \\
\dfrac{dM_{tetR}}{dt} = -M_{tetR} + \dfrac{10.338}{1.035+P_{cI}^{3}}\vspace{1ex} \\
\dfrac{dM_{cI}}{dt} = -M_{cI} + \dfrac{9.845}{0.987+P_{lacI}^{3}}\vspace{1ex} \\
\dfrac{dP_{cI}}{dt} = M_{lacI} - P_{cI}\vspace{1ex} \\
\dfrac{dP_{lacI}}{dt} = M_{tetR} - P_{lacI}\vspace{1ex} \\
\dfrac{dP_{tetR}}{dt} = M_{cI} - P_{tetR}\vspace{1ex}, \\
\end{cases}
\label{eq:protein-my}
\end{equation}

%%baselines
\section{Baselines on Gene Regulatory Networks}\label{app:baseline_bio}
We also adopt the baseline methods to identify the governing equations of two gene regulatory networks. As illustrated in Table~\ref{tab:baseline-bio}, we can observe that the governing equations uncovered by our ScaleSR method are close to the ground truth, while the baselines fail to discover governing equations from data. Note that we only list one representative equation in the following Table, but you can refer to the discovered equations of our method in Eqs.\ref{eq:protein-my} and \ref{eq:toggle-my} above. Thus, the proposed ScaleSR demonstrates the superior performance over the baselines in discovering symbolic equations.

\begin{table}[!thb]
\centering
\caption{The discovered results of different methods (one representative equation)}
\label{tab:baseline-bio}
\begin{adjustbox}{width=0.95\textwidth}
\begin{tabular}{lll}
\hline
\textbf{Task} & \textbf{Model} & \textbf{Discovered Equation} \\ 
\hline
Toggle & Truth & $dU / dt = 4 / (1+V^{3})-U$ \\ 
 & ScaleSR & $dU / dt = 3.919 / (0.972+V^{3})-U$ \\ 
 & SPL & $dU / dt = -U + 2.539 \exp(\cos(V)) - \cos(V) - 1.749$ \\
 & NGGP & $dU / dt = U - V + \cos(V - \log(-1 / (0.084 U^2) + 0.084 U \log(U) - 3.194)) + 0.067) + 3.544$ \\
 & DSR & $dU / dt = -U - V - sin(1.357 V + 5.138) + 3.357$ \\
 & GP & $dU / dt = \sin(\exp(\exp(\exp(U))^2) - \exp(\exp(\sin(\exp(V))^2 \cdot \sin(\cos(U) - U - 7.252))$  \dots\\ 
\hline
Protein & Truth & $dM_{lacI} / dt = -M_{lacI} + 10 / (1+P_{tetR}^{3}) + 10^{-5}$ \\
 & ScaleSR & $dM_{lacI} / dt = -M_{lacI} + 9.939 / (0.982+P_{tetR}^{3})$ \\ 
 & SPL & $dM_{lacI} / dt = 3.234 - 0.455 M_{tetR}$ \\
 & NGGP & $dM_{lacI} / dt = M_{lacI}(-2 P_{cI} P_{tetR} +P_{lacI} + 6.564)/(P_{cI} + 10.913)$ \\
 & DSR & $dM_{lacI} / dt = -1.582 + 3.482 M_{tetR}^2 / (P_{tetR} M_{tetR}^2 + 0.842)$ \\
 & GP & $dM_{lacI} / dt = 2.162 / (M_{cI}^7 + M_{cI} P_{lacI})$ \\ 
 \hline

\end{tabular}
\end{adjustbox}
\end{table}

\section{Limitations and Future Work}
The accuracy of our evaluation results is impacted by the accuracy of single-variable symbolic regression. In this work, we adopt the state-of-the-art MCTS method for symbolic regression. For future work, we need to develop new single-variable symbolic regression models to improve the accuracy. The accuracy of prediction results is also impacted by the number of training samples for DNNs. If the number of training data is limited, we need to explore a physics-enhanced neural symbolic regression model.

% In addition, the data noise will influence the accuracy of symbolic regression. We will explore a more robust method to reduce noise sensitivity.

% Finally, the accuracy of prediction results is also impacted by the accuracy of DNNs.

\section{Broader Impact}
The goal of this work is to improve the accuracy and scalability of symbolic regression for scientific discovery. The proposed ScaleSR method has demonstrated superior performance over state-of-the-art methods in discovering analytical expressions from data, which can promote AI for scientific discovery. Note that this fundamental research will not cause any potential negative societal impacts.

\section{Computing Resources}
We implement our experiments on the server with 1 A5000 GPUs with 24 GB graphics memory. The server has 32-Core 3.4 GHz AMD EPYC 7532 processors with 250GB RAM with 4TB SSD of storage capacity.

% \section{Temporary Section: Figs}
% \begin{figure}
%     \centering
%     \includegraphics[width = 8cm]{fig_hongjue/toggle.pdf}
%     \caption{Toggle Switch}
%     \label{fig:hongjue_toggle}
% \end{figure}

% \begin{figure}
%     \centering
%     \includegraphics[width = 12cm]{fig_hongjue/Repressilator.pdf}
%     \caption{Repressilator}
%     \label{fig:hongjue_Repressilator}
% \end{figure}

% \begin{figure}
%     \centering
%     \includegraphics[width = 8cm]{fig_hongjue/protein.pdf}
%     \caption{protein}
%     \label{fig:hongjue_protein}
% \end{figure}

\end{document}